\pdfoutput=1
\documentclass[11pt]{article}
\usepackage[margin=1in]{geometry}
\usepackage[T1]{fontenc}
\usepackage[utf8]{inputenc}
\usepackage{lmodern}
\usepackage{microtype}
\usepackage{amsmath,amssymb}
\usepackage{booktabs}
\usepackage{tabularx}
\usepackage{array}
\usepackage{graphicx}
\usepackage{caption}
\usepackage{float}
\usepackage{placeins}
\usepackage{titlesec}
\usepackage{url}
\usepackage[hidelinks]{hyperref}
\usepackage{textcomp}
\begin{document}

\title{Measuring Asset and Scene Reconstruction Effects in Real-to-Sim Robot Evaluation}
\author{Sanya Verma, Luca Cilio, Velissarios Christodoulou, Tyler Fermelis, \\ Qi Ting Ng, Rachel Chung, Sofia Liang \\[3pt] Kaedim}
\date{}
\maketitle

\begin{abstract}
Simulated evaluation is increasingly used alongside real-world evaluation of robot policies because it is cheaper and easier to repeat; however, its value depends on how closely its outcomes track the real robot's. We test whether our reconstruction pipeline, combining metrically scaled object geometry, authored physical parameters and scene reconstruction, reduces disagreement between simulated and real robot scores relative to a default open-source recipe. We constructed two simulated versions of one bimanual robot cell: an authored reconstruction, using object geometry at estimated metric scale, projected textures, authored physics and our own scene splat; and a baseline, referred to as the default reconstruction, using the open-source recipe of a generative single-image mesh, engine-default physics and a Gaussian-splat scene. Both reconstructions use the same object photographs and scene video. Two policies ran five tasks each, giving ten task-policy pairs, which we call cells; each cell was run twenty times in each reconstruction with all other settings held fixed. Both reconstructions were scored against the same real trials, graded by a third-party evaluator. Pearson correlation between the ten simulated and real cell means is r = 0.90 for the authored reconstruction and 0.51 for the default. Mean score error is 6.97 percentage points for the authored reconstruction and 17.54 for the default, a reduction of 10.56 percentage points. These results show that improving the quality of the environment reconstruction through higher visual fidelity, authored physics and metric scale makes the simulation more faithful to the real world and narrows the sim-to-real gap. We release the harness, the per-trial scores, every reported run's configuration, and the assets and scenes of both reconstructions.

\end{abstract}

\FloatBarrier
\section{Introduction}\label{sec:1}
Evaluating a robot policy on hardware is costly and labour-intensive, and because real rollouts can be stochastic and hard to control systematically, a result is difficult to repeat exactly \cite{ref1,ref2}. One alternative to hardware is a digital twin: a simulated copy of the real robot cell in which the same policy runs the same tasks. SIMPLER \cite{ref1} reported a strong correlation between simulated and real performance for the policies and tasks it studied. PolaRiS \cite{ref2} and SimFoundry \cite{ref3} build the twin from a short video scan of the robot cell and report Pearson correlations (r) of 0.90 and 0.91 respectively between simulated and real scores.

The remaining disagreement between real and simulated scores is attributed to two sources, both identified by SIMPLER \cite{ref1}. The first is the visual gap: rendered images differ from real camera images. The second is the control gap: the simulated arm responds differently to the same command. Later work varies factors within these two categories. PolaRiS \cite{ref2} compares four rendering conditions of the same scene, from Gaussian splatting to fully untextured, and measures the change in r and mean maximum rank violation (MMRV). Wang et al. \cite{ref4} vary vision, language, layout and behaviour across nine tasks and five policies. How the manipulated objects themselves are reconstructed is not among the sources these studies vary.

In these pipelines, object physics is not investigated as a source of disagreement between real and simulated scores: it is left at engine defaults or estimated by a model without validation against measurement. PolaRiS builds object geometry with open-source TRELLIS \cite{ref9} from several segmented views of each object, then sets ``standard physics parameters for contact dynamics using default values from IsaacSim'' and notes that ``object mass parameters can be estimated by the users'' \cite[Section~\ref{sec:4}.1]{ref2}; object physics is left at engine defaults. SimFoundry \cite{ref3}, like our pipeline, authors mass and friction per object, but reports only how well its complete output agrees with real, so the effect of that step is not measured.

Prior results suggest that the effect of authored physics is not negligible. Phys2Real \cite{ref5} shows that better estimates of an object's centre of mass raise real task success, although it measures success, not agreement between simulated and real evaluation. The nearest evidence on agreement between real and simulated evaluation is Wang et al. \cite{ref4}, who find that agreement is governed by simulator fidelity instead of purely object-level mismatch; their test varies which object is present, not how well a fixed object is built. The only measurement of asset fidelity against agreement is on deformable objects: Zhang et al. \cite{ref6} replace PhysTwin's \cite{ref8} optimised per-spring stiffness with a single global value and MMRV on a rope-routing task rises from 0.174 to 0.248. No study that we are aware of measures how the fidelity of a fixed set of rigid objects changes agreement between simulated and real evaluation (Section~\ref{sec:2}). We therefore build one cell twice: an authored reconstruction, with our pipeline, which reconstructs geometry from one photograph, estimates metric scale from a ChArUco calibration board and authors physics per object, and a default reconstruction, a PolaRiS-inspired recipe, TRELLIS at its single-image setting with engine-default physics (Section~\ref{sec:3}).

We test whether the authored reconstruction narrows the sim-to-real gap in robot performance relative to the default reconstruction. Both reconstructions are run with the same policies and tasks and scored against the same real trials (Figure~\ref{fig:2}). The reconstructions differ on three axes. Object geometry and texture: reconstructed and projected from a single photograph of the object, scaled from a printed ChArUco calibration board of known size (authored), against a generative single-image mesh scaled by eye (default). Object physics: mass, friction and restitution authored per object, against engine defaults. Scene: our reconstruction splat pipeline, against a PolaRiS-inspired Gaussian-splat recipe, both from the same handheld video. Every other part of the run (robot model, controller, camera, placement and step budget) is identical between the reconstructions (Section~\ref{sec:3}).

Our main finding is that a reconstruction with metrically scaled geometry, authored physics and our own scene reconstruction taken together narrows the sim-to-real gap relative to the default recipe used in the literature: score error is 10.56 percentage points lower and progress disagreement 8.32 percentage points lower across five tasks and two policies, both with 95\% intervals that exclude zero. The other two agreement measures, correlation and failure-stage disagreement, do not separate the two reconstructions with ten cells. We release the harness, the four agreement measures and the real reference of 200 paired trials, five tasks by two policies, run and graded by a third-party evaluator \cite{ref11}, and the assets and scenes of both reconstructions, so the comparison can be repeated on other cells and pipelines.

\FloatBarrier
\section{Related work}\label{sec:2}
Prior work relevant to this study falls into three groups: real-to-sim evaluation systems, the pipelines those systems use to create assets, and methods that estimate physical properties. 

For real-to-sim evaluation systems, SIMPLER \cite{ref1} reported a strong correlation between simulated and real performance for the policies and tasks it studied, named the visual and control gaps, and introduced mean maximum rank violation (MMRV), which measures whether a simulator ranks policies in the same order as the real robot. PolaRiS \cite{ref2} reconstructs the robot cell from a video scan and reports r of 0.9 on average and 0.81 at worst over six environments and four policies. Its ablation of the environment itself is visual: the same scene is rendered in four ways, as a Gaussian splat, as a ray-traced textured mesh, with the background untextured, and fully untextured, and both r and MMRV worsen as texture is removed. SimFoundry \cite{ref3} reports r = 0.911 and MMRV 0.018 over seven tasks and five policies, and a higher correlation than a PolaRiS baseline on its own tasks; it varies nothing about the assets. Wang et al. \cite{ref4} extend to nine tasks and five policies and argue that a reliable simulator must also reproduce failure modes. Their one test on objects replaces the manipulated object in a pick-and-place task (a banana replaced by five other produce items) and finds the policy ranking unchanged, from which they conclude that simulator fidelity, not the choice of object, drives the gap between real and simulated performance. Each of these studies builds one simulated copy of each real setup and reports how well that copy agrees with reality; none builds the same setup twice and measures how the agreement changes.

Among pipelines that create assets, PolaRiS \cite{ref2} leaves mass unset unless the user enters it, so PhysX assigns a density of 1000 kg/m$^{3}$ over the collider volume, friction 0.5 and restitution 0.0. Its published object assets carry a single uniform root scale and convex-decomposition colliders. SimFoundry \cite{ref3} is the pipeline closest to ours in what it automates: from an RGB video it segments objects, generates meshes with an image-to-3D model, aligns them to the scene's RGB-D reconstruction, builds CoACD convex colliders, and assigns mass and friction per object from a scene-understanding vision-language model. It differs from our authored reconstruction in geometry (a different image-to-3D model), scale (from depth, not a ChArUco board), collision (convex decomposition throughout, against signed distance fields on seven of our sixteen object kinds) and physics (per object from a vision-language model in both; ours combines the model's class, mass and restitution with a per-class friction table). 

The third group estimates physical properties of objects. Phys2Real \cite{ref5} combines vision-language priors over an object's centre of mass with online identification from interaction, and finds that better estimates raise real task success. TwinAligner \cite{ref7} optimises mass, friction and centre of mass against real trajectories. Both methods improve the estimate of an object's physical properties, but neither measures how the quality of the estimate changes agreement between simulated and real evaluation. The nearest such measurement is on deformable objects: Zhang et al. \cite{ref6} remove PhysTwin's \cite{ref8} per-spring stiffness optimisation, leaving a single global value, and MMRV on a rope-routing task rises from 0.174 to 0.248. No study we are aware of varies the geometry, scale or physics of a fixed set of rigid objects and measures the change in agreement.

\FloatBarrier
\section{Experimental setup}\label{sec:3}
The robot setup is a bimanual I2RT YAM station \cite{ref10}. This study covers five tasks and two policies, giving ten cells; each cell is run twenty times in each reconstruction, so 200 trials per reconstruction and 400 in all, paired against 200 real trials. The tasks are placing bottles in a bin, stacking bowls, stacking blocks, moving a latte cup onto a board, and clearing a table into a bin. These tasks use twenty-one objects: six bottles of three different kinds and a bin; three identical bowls; two blocks; a cup, can, spray bottle and cutting board; and a cube, baseball, bunch of grapes, mug and plate. Stacking bowls and stacking blocks are the evaluator's standard tasks; the other three were specified for this study. Moving a latte cup onto a board follows the PolaRiS task of the same name \cite{ref2} and clearing a table follows SimFoundry's \cite{ref3}; the two stacking tasks have counterparts in both, block stacking in PolaRiS and dishware stacking in SimFoundry. Objects, robot setup, rubrics and step budgets were supplied by the evaluator. 

The real trials were run and graded by Robocurve, a third-party robot evaluator, on its own robot setup under its own protocol. Kaedim specified the three custom tasks and agreed the policies with Robocurve; Robocurve wrote the rubrics and, under the benchmarking agreement that commissioned the trials, retained control of methodology, scoring and interpretation on the real side. Kaedim, the authors' employer, paid for the trials (see Funding and competing interests). Per-trial scores, rubrics, step budgets and rig assignments (which of Robocurve's robot arm stations ran a trial) are published by Robocurve \cite{ref11}; the scores are also released with this paper. The rubrics award credit at stages: for a bottle, 5 points for grasped and lifted, 8 for in the bin on its side and 10 for upright. The full rubric is given with the grading method in Section~\ref{sec:4}.

The policies are $\pi$0.5 \cite{ref12} and MolmoAct2 \cite{ref13}. The evaluator runs $\pi$0.5 with the top camera 0.72 m above the table and MolmoAct2 at 0.88 m \cite{ref11}; the simulation matches each.

Table~\ref{tab:1} summarises the difference between the authored and default reconstructions. Everything not listed is identical between them.

\begin{table}[htbp]
\centering
\small
\setlength{\tabcolsep}{4pt}
\renewcommand{\arraystretch}{1.15}
\caption{The two reconstructions.}
\label{tab:1}
\begin{tabularx}{\textwidth}{>{\raggedright\arraybackslash}X>{\raggedright\arraybackslash}X>{\raggedright\arraybackslash}X}
\toprule
 & \textbf{authored reconstruction} & \textbf{default reconstruction} \\
\midrule
object geometry & mesh reconstructed with our pipeline from a photograph & mesh generated by TRELLIS from the same photograph \\
texture & projected from the same photograph & generated by TRELLIS with the mesh \\
metric scale & estimated by our pipeline from a printed ChArUco board of known size captured with the object & typed in by hand to match the object's apparent size in the same photograph; no real measurement \\
collision geometry & signed distance field (bowl, cup, can, spray bottle, cube, mug, plate); authored convex hulls (bottles, blocks, cutting board); bounding sphere (baseball); convex decomposition (grapes, bin) & convex decomposition \\
mass, friction, restitution & mass and restitution estimated by a vision-language model per object; friction from a per-class table; material class from the same model & not authored; PhysX defaults, density 1000 kg/m$^{3}$, friction 0.5, restitution 0 \\
scene reconstruction, from the same handheld video & 3D Gaussian splat \cite{ref14} trained from the handheld video, aligned by a table-plane fit, cut above the table & 2D Gaussian splat \cite{ref16} by a PolaRiS-inspired recipe from the same video, same alignment and cut \\
\bottomrule
\end{tabularx}
\end{table}
\FloatBarrier

The authored reconstruction's objects are built with our asset pipeline (\href{https://www.projectsim.ai}{projectsim}); its scene is built with our own scene reconstruction pipeline. Geometry is reconstructed from a single photograph, and metric scale is estimated from a printed ChArUco board of known size captured with the object. Colliders are signed distance fields on the bowl, cup, can, spray bottle, cube, mug and plate, authored convex hulls on the bottles, blocks and cutting board, a bounding sphere on the baseball, and convex decompositions on the grapes and the bin. A vision-language model assigns each object a material class and estimates its mass and restitution from rendered views of the reconstructed mesh; static and dynamic friction are taken from a per-class table keyed on that class. 

\begin{figure}[htbp]
\centering
\includegraphics[width=\linewidth]{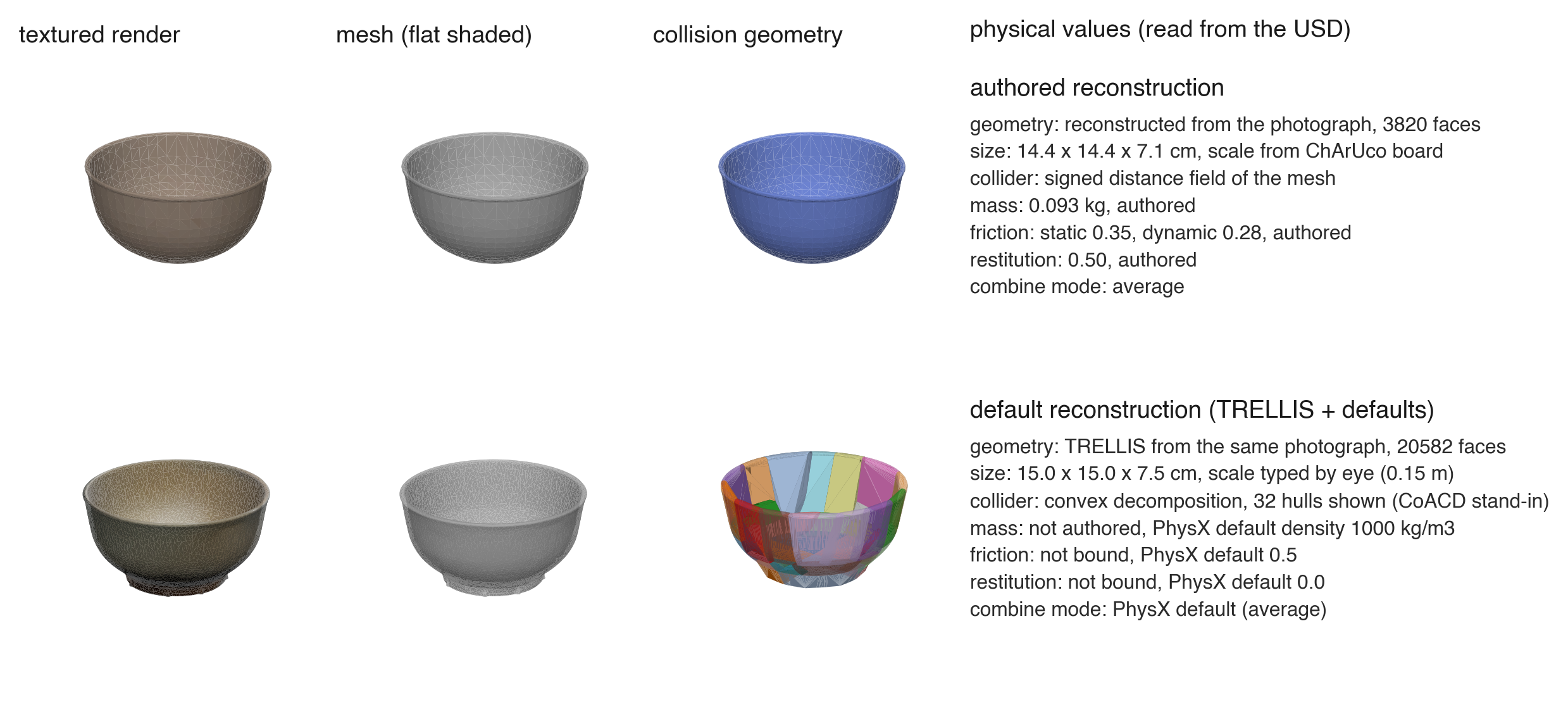}
\includegraphics[width=\linewidth]{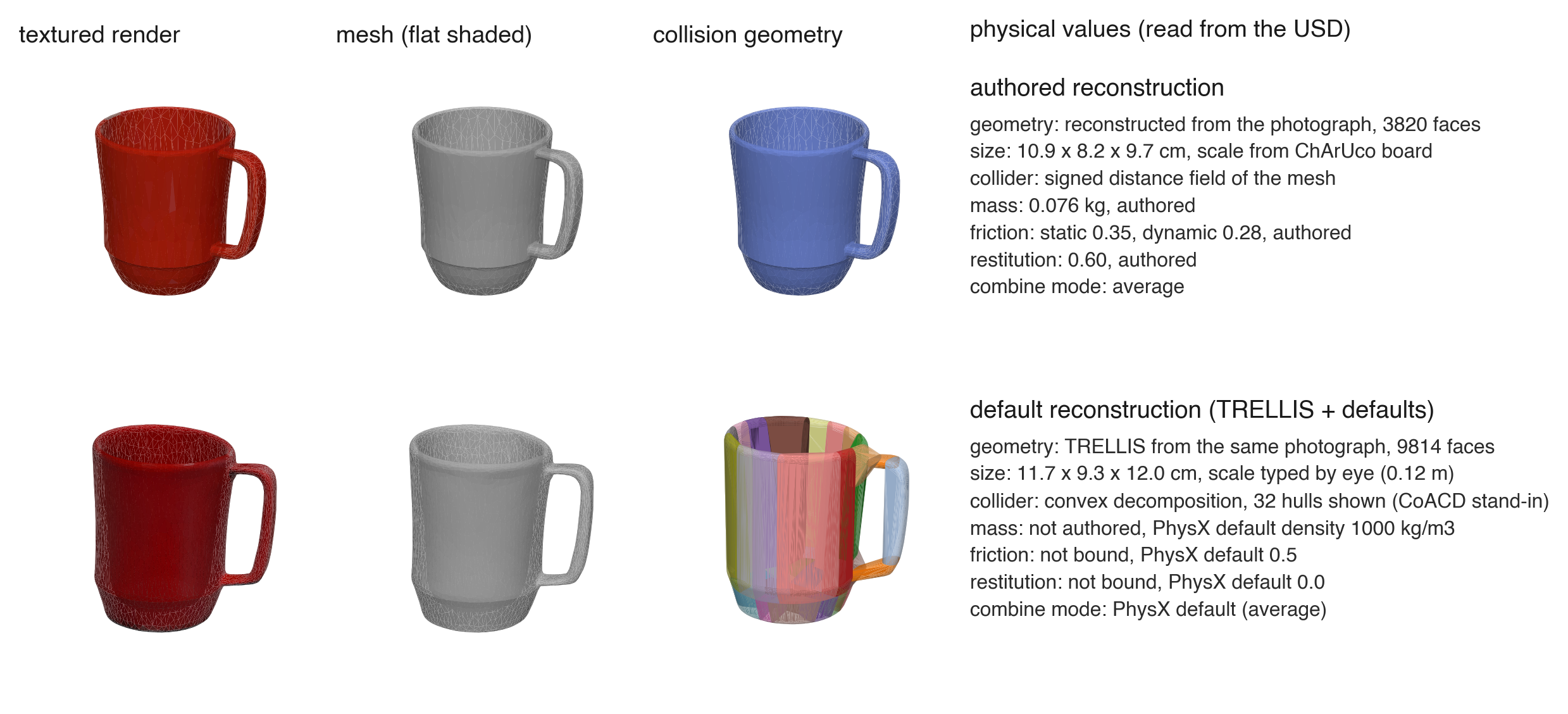}
\caption{Two objects, the ribbed bowl (top) and the red mug (bottom), in both reconstructions. Per reconstruction, left to right: textured render, mesh (flat shaded), what PhysX collides with, and the values read from each USD; the authored objects are a signed distance field of their own mesh, the default objects are a convex decomposition that fills the bowl's cavity and the mug's interior (32 hulls each, CoACD \cite{ref15} as a stand-in for the decomposition PhysX computes at load). All panels of an object share one camera.}
\label{fig:1}
\end{figure}

The default reconstruction follows a PolaRiS-inspired recipe \cite{ref2}. A single photograph of each object, one of the same capture photographs we use for the authored reconstruction, goes through TRELLIS \cite{ref9}, which generates both mesh and texture. Metric scale is typed by hand from the photograph and everyday familiarity with the object, with no measurement used; PolaRiS likewise leaves scale to the user's discretion. Collision geometry is a convex decomposition, which fills concave shapes such as bowls and bins. No physics is authored, so PhysX defaults apply: friction 0.5, restitution 0, and mass from a density of 1000 kg/m$^{3}$ over the collider volume. The scene is a 2D Gaussian splat trained from the same handheld input video by the PolaRiS-inspired recipe, COLMAP poses \cite{ref17} followed by 2D Gaussian Splatting \cite{ref16}, then aligned and cut exactly as the authored scene (below). We built this reconstruction ourselves from the published recipe. Figure~\ref{fig:1} shows two objects under both reconstructions.

\FloatBarrier

Two inputs differ between the reconstructions: the object assets and the scene splat. Every other setting is held fixed and checked from the run logs (Section~\ref{sec:4}). The robot model is the same file in both reconstructions; only the objects and the scene are reconstructed. Both reconstructions take the same inputs: the object photographs, and a single handheld video of the setup from which both scene splats were trained. A trained splat has no metric scale, orientation or position of its own. Orientation comes from the floor: the largest flat surface in the splat is the floor, and the splat is rotated so that this surface is horizontally level. Scale is assigned using the table: the height band where most splat points lie is the table top, and matching its size to the real table gives 0.807 m per splat unit. Position was then set by hand so that the splat's table top sits at the measured 0.750 m of the real table and the arm mounts sit on its near edge. The input video also captured the objects on the table, and rendering them would show every object twice, once from the splat and once from the simulator, so both splats are cut to remove everything above the table surface. The default splat is registered onto the authored one after alignment, so both share the same frame and the same cut.

\begin{figure}[htbp]
\centering
\includegraphics[width=\linewidth]{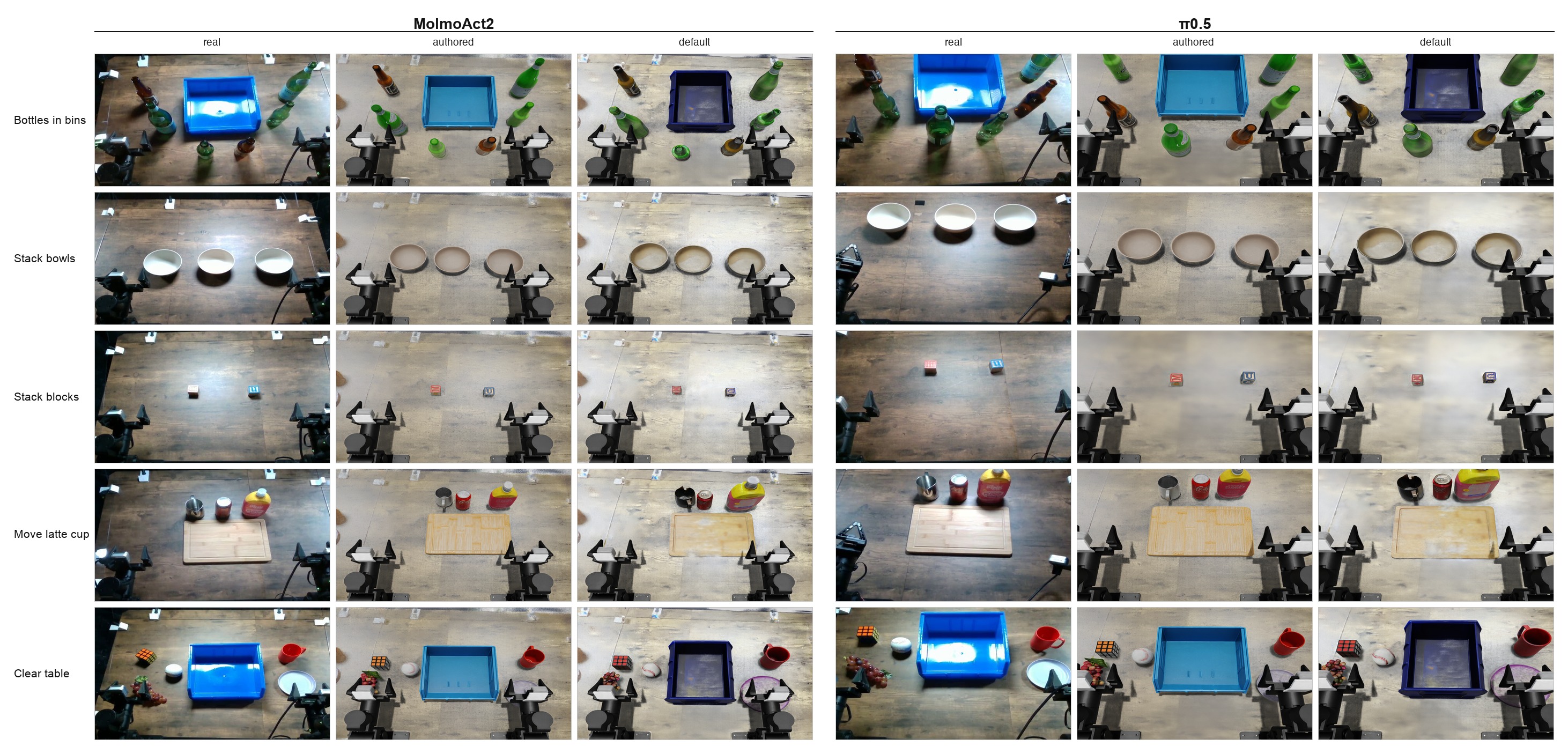}
\caption{First top-camera frame of each of the ten cells, real against both reconstructions.}
\label{fig:2}
\end{figure}

Objects start each trial at fixed positions per cell, taken from one real trial's first camera frame. Each of the twenty trials then shifts every object by up to 1 cm across the table, so that the trials do not all start from one identical placement. Real placement varies more than this from trial to trial, so simulated placement is narrower than real. Placement is identical between the authored and default reconstructions, so it does not contribute to the difference between them. Real and simulated trials are therefore paired by cell, not by trial. Figure~\ref{fig:2} shows the first top-camera frame of every cell.

\FloatBarrier

\FloatBarrier
\section{Method}\label{sec:4}
The harness takes four inputs: the robot model, the tasks with their real per-trial scores, a policy served as a separate process, and a reconstruction (objects and scene). The first three are fixed for this study; only the reconstruction changes. The harness runs the policy in Isaac Sim 6.0.1 for twenty trials per cell and writes a per-trial score under the evaluator's rubric.

A configuration gate decides which runs may be compared, so that a difference between reconstructions cannot come from a setting that drifted between runs. Nine properties must match: step budget, action chunk length, top camera height, placement version and its per-policy key, physical rig, gripper collision filter, joint clipping and solver step count. Table~\ref{tab:2} lists the settings held constant across both reconstructions. A run that fails any check is rejected with the reason printed. During this study the gate rejected six complete default-reconstruction runs that had rendered at the wrong camera height, a defect that leaves no trace in the scores. The gate's limit is that it sees only what the runner prints; the per-trial placement shift, for example, is not logged.

\begin{table}[htbp]
\centering
\small
\setlength{\tabcolsep}{4pt}
\renewcommand{\arraystretch}{1.15}
\caption{Settings held constant, each read from the reported runs' logs and job records.}
\label{tab:2}
\begin{tabularx}{\textwidth}{>{\raggedright\arraybackslash}X>{\raggedright\arraybackslash}X}
\toprule
\textbf{setting} & \textbf{value} \\
\midrule
control & 30 Hz; absolute joint targets clipped to each joint's range, as on the evaluator's controller \\
action chunk & 16 steps per inference ($\pi$0.5), 30 (MolmoAct2) \\
MolmoAct2 solver steps & 10 \\
step budget per trial & bottles 6400, bowls 3600, blocks 3600, latte 3600, clear table 12800, matching the evaluator \\
top camera & rendered 640x480, served at 640x360; 70 degrees below horizontal; 0.72 m ($\pi$0.5) or 0.88 m (MolmoAct2); measured D435 intrinsics per rig \\
object placement & one layout per cell refit from the evaluator's start frame; up to 1 cm seeded jitter in x and y per trial \\
physics & PhysX 240 Hz; continuous collision detection on; contact offset 6 mm; 64 position and 16 velocity solver iterations \\
gripper & finger-to-finger collision filtered; drive force 100 N; pads unbound, so PhysX default friction 0.5 is averaged with the object's \\
table surface & static friction 0.55, dynamic 0.45, restitution 0.02 \\
trials & 20 per cell per reconstruction; 200 per reconstruction, 400 in all \\
\bottomrule
\end{tabularx}
\end{table}
\FloatBarrier

Once a run passes the gate, its trials are scored. Real trials were graded by the evaluator's team, against its rubric for every task (Table~\ref{tab:3}). Simulated trials in both reconstructions were graded automatically by rules derived from the same rubric, with each stage decided by a rule on object poses and gripper contact, so the two reconstructions are scored identically; the rules are part of the released harness.

\begin{table}[htbp]
\centering
\small
\setlength{\tabcolsep}{4pt}
\renewcommand{\arraystretch}{1.15}
\caption{Rubric per stage. Task strings and points are the evaluator's \cite{ref11}.}
\label{tab:3}
\begin{tabularx}{\textwidth}{>{\raggedright\arraybackslash}X>{\raggedright\arraybackslash}X>{\raggedright\arraybackslash}X>{\raggedright\arraybackslash}X}
\toprule
\textbf{task (max)} & \textbf{task string} & \textbf{stage} & \textbf{points} \\
\midrule
bottles in bin (20) & ``put 2 glass bottles in the blue bin'' & grasped and lifted & 5 \\
 &  & in bin, on its side & 8 \\
 &  & in bin, upright & 10 \\
 &  & total score of two bottles in bin & 20 \\
stack blocks (30) & ``stack the red block on the blue block'' & attempt & 5 \\
 &  & red lifted & 10 \\
 &  & red carried over blue & 20 \\
 &  & red resting on blue & 30 \\
stack bowls (50) & ``stack the bowls on the sides on top of the center bowl'' & bowl lifted & 5 each \\
 &  & two bowls almost stacked & 20 \\
 &  & two bowls stacked & 30 \\
 &  & three almost stacked & 40 \\
 &  & three stacked & 50 \\
move latte cup (10) & ``move the steel latte art cup onto the cutting board'' & moved & 5 \\
 &  & on board & 8 \\
 &  & on board, upright & 10 \\
clear table (70) & ``put all items on the table inside the blue bin'' & moved / lifted / in bin, per item & cube 5/10/15, baseball 5/10/15, plate 10/17/20, mug 5/8/10, grapes 5/8/10 \\
\bottomrule
\end{tabularx}
\end{table}
\FloatBarrier

We report four measures of agreement between a reconstruction and the real robot, all computed on scores expressed as a percentage of the task's maximum (10, 20, 30, 50 or 70 points, Table~\ref{tab:3}) so that tasks with different point scales can be averaged. Two measures compare the score numbers themselves, score error and correlation; two compare the stage at which trials stop, progress and failure-stage disagreement.

Let c = 1, \ldots{}, 10 index the cells, let $x_{c,i}$ and $y_{c,j}$ (i, j = 1, \ldots{}, 20) be the simulated and real trial scores in cell c, and let $\bar{x}_c$ and $\bar{y}_c$ be their means over the twenty trials.

Score error measures how far the simulated score is from the real score. It is the absolute difference between the simulated and real cell means, averaged over the ten cells:

\begin{equation}
E=\frac{1}{10}\sum_{c=1}^{10}\left|\bar{x}_c-\bar{y}_c\right|
\end{equation}
Correlation measures whether cells that score high in reality also score high in simulation. It is Pearson's correlation r between the ten simulated and the ten real cell means, with $\bar{x}$ and $\bar{y}$ the means of those cell means:

\begin{equation}
r=\frac{\sum_{c}(\bar{x}_c-\bar{x})(\bar{y}_c-\bar{y})}{\sqrt{\sum_{c}(\bar{x}_c-\bar{x})^2\;\sum_{c}(\bar{y}_c-\bar{y})^2}}
\end{equation}
Progress disagreement measures whether the robot gets as far through the task in simulation as in reality. Let $R_c$ be the rubric stages the real robot reached at least once in cell c, specifically its distinct non-zero scores. For each stage, the fraction of simulated trials that reached it is compared with the real fraction; the absolute gaps between the simulated and real fractions are averaged over stages, then over cells:

\begin{equation}
P=\frac{1}{10}\sum_{c=1}^{10}P_c,\qquad P_c=\frac{100}{|R_c|}\sum_{\rho\in R_c}\left|\frac{1}{20}\sum_{i=1}^{20}\mathbf{1}[x_{c,i}\ge\rho]-\frac{1}{20}\sum_{j=1}^{20}\mathbf{1}[y_{c,j}\ge\rho]\right|
\end{equation}
where $\mathbf{1}[\cdot]$ is 1 when the condition holds and 0 otherwise. For example, if 60\% of real trials and 40\% of simulated trials reached the ``in bin'' stage, that stage contributes a gap of 20 percentage points.

Failure-stage disagreement measures whether trials stop at the same stage. Sort the stages of $R_c$ as $\rho_1<\dots<\rho_{|R_c|}$, with $\rho_0=0$ and $\rho_{|R_c|+1}=\infty$, and let $p^{x}_{c,k}$ be the fraction of simulated trials in cell c whose score lies in $[\rho_k,\rho_{k+1})$, that is, that stopped at stage k, and $p^{y}_{c,k}$ the same for real trials. The measure is the total variation distance between the two distributions, averaged over cells:

\begin{equation}
F=\frac{1}{10}\sum_{c=1}^{10}F_c,\qquad F_c=100\cdot\frac{1}{2}\sum_{k=0}^{|R_c|}\left|p^{x}_{c,k}-p^{y}_{c,k}\right|
\end{equation}
Correlation is higher when simulation is closer to reality; the other three are lower.

For uncertainty, every comparison is paired on the cell, so that the large differences in difficulty between tasks and cells cancel. All intervals are bootstrap percentile intervals: the statistic is recomputed on each of 20,000 resamples and the 2.5th and 97.5th percentiles give the 95\% interval. A cell is considered consistent with real when the interval on the difference between its simulated and real means includes zero. For each resample b, twenty simulated and twenty real scores are drawn with replacement from the cell, and

\begin{equation}
D_c^{(b)}=\bar{x}_c^{(b)}-\bar{y}_c^{(b)},\qquad \text{cell } c \text{ is consistent with real if } 0\in\left[Q_{2.5}(D_c),\,Q_{97.5}(D_c)\right]
\end{equation}
where $Q_p$ is the p-th percentile over resamples. This is the only test used when a cell is called consistent or inconsistent with real. To compare the two reconstructions on E, P and F, let $M_c$ be the per-cell value of the measure and $d_c = M_c^{\mathrm{default}} - M_c^{\mathrm{authored}}$, so that a positive $d_c$ means the authored reconstruction is closer to real in cell c. Each resample draws ten cells with replacement and takes their mean difference,

\begin{equation}
\bar{d}^{(b)}=\frac{1}{10}\sum_{c\in S^{(b)}}d_c,\qquad \text{95\% interval}=\left[Q_{2.5}(\bar{d}),\,Q_{97.5}(\bar{d})\right]
\end{equation}
where $S^{(b)}$ is the ten cells drawn. For correlation, each resample draws ten cells, recomputes r for both reconstructions on them, and takes the difference; the interval is the same percentiles of those differences.

A batch is twenty trials of one cell under one set of settings, from a single run or from consecutive partial runs pooled in order. Where more than one batch passed the gate for a cell and reconstruction, we report only the most recent by the run's own logged start time. Every passing batch is listed in Appendix~\ref{app:A}, and the reported run list is released (Appendix~\ref{app:C}).

\FloatBarrier
\section{Results}\label{sec:5}
Table~\ref{tab:4} gives the four agreement measures for both reconstructions and the paired cell-by-cell difference between them. Score error is 6.97 percentage points for the authored reconstruction and 17.54 for the default, a difference of 10.56 percentage points (95\% CI 5.03 to 17.11). Progress disagreement is 15.90 percentage points for the authored reconstruction and 24.23 for the default, a difference of 8.32 percentage points (1.78 to 16.74). For both measures, differences are statistically significant: the 95\% bootstrap intervals exclude zero (Table~\ref{tab:4}). Correlation with the real cell means is r = 0.90 (0.77 to 0.99) for the authored reconstruction and 0.51 (0.02 to 0.94) for the default; failure-stage disagreement is 42.00 for the authored reconstruction and 48.50 for the default. On these two measures the intervals on the difference include zero, so with ten cells they do not separate the reconstructions.

\begin{table}[htbp]
\centering
\small
\setlength{\tabcolsep}{4pt}
\renewcommand{\arraystretch}{1.15}
\caption{The four agreement measures, paired on the cell. Correlation is higher when closer to real; the three disagreement measures are lower. A positive difference means the authored reconstruction is closer to real.}
\label{tab:4}
\begin{tabularx}{\textwidth}{>{\raggedright\arraybackslash}Xll>{\raggedright\arraybackslash}Xl}
\toprule
\textbf{measure} & \textbf{authored} & \textbf{default} & \textbf{paired difference, 95\% CI} & \textbf{excludes zero} \\
\midrule
score error (percentage points) & 6.97 & 17.54 & +10.56 [+5.03, +17.11] & yes \\
progress disagreement (percentage points) & 15.90 & 24.23 & +8.32 [+1.78, +16.74] & yes \\
Pearson correlation (r) & 0.90 & 0.51 & +0.39 [-0.08, +0.87] & no \\
failure-stage disagreement (percentage points) & 42.00 & 48.50 & +6.50 [-3.50, +17.00] & no \\
\bottomrule
\end{tabularx}
\end{table}
\FloatBarrier

\begin{figure}[htbp]
\centering
\includegraphics[width=0.78\linewidth]{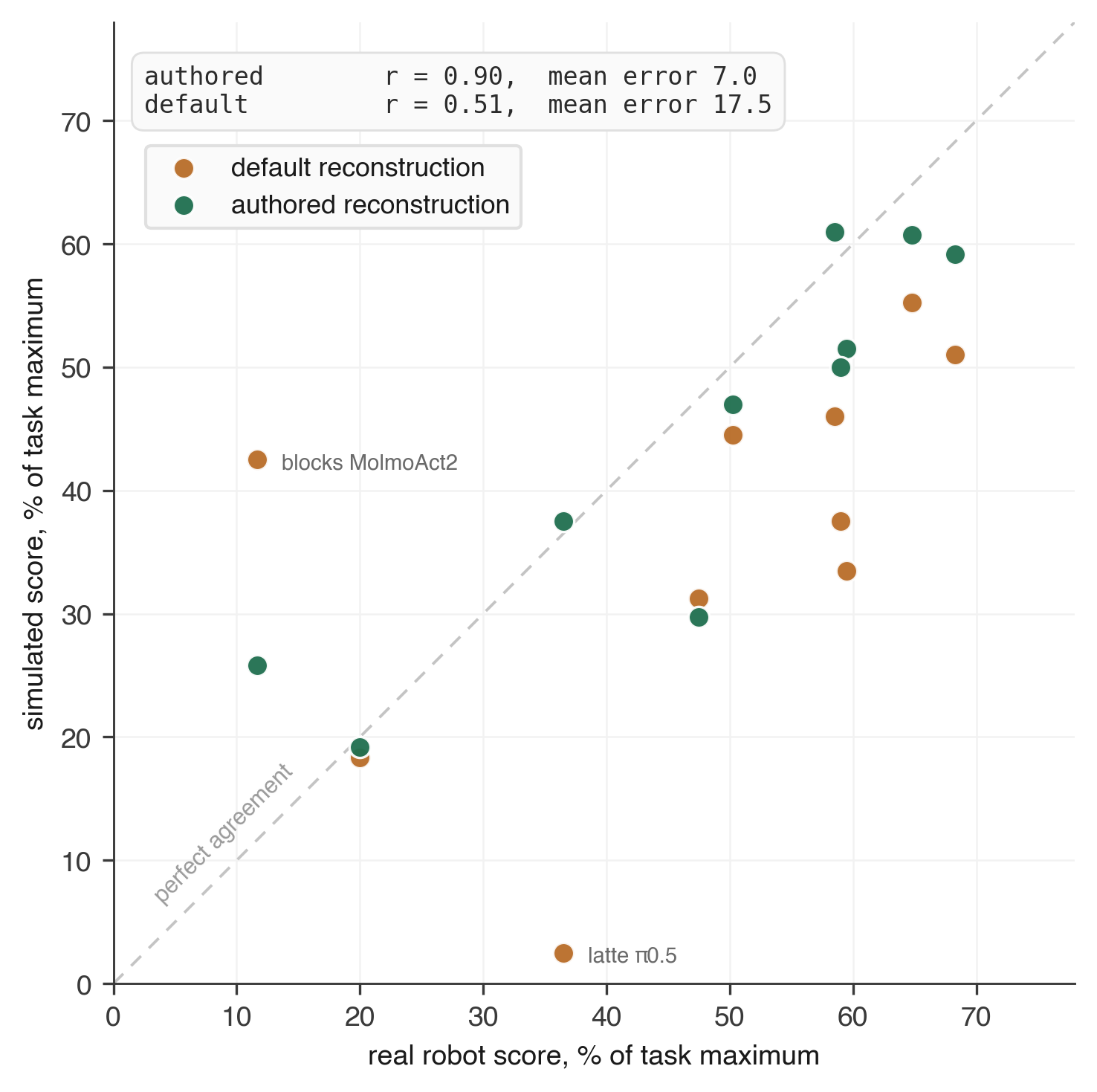}
\caption{Simulated against real mean score per cell, percent of task maximum. Each point is one cell; the dashed line is perfect agreement. The two labelled points are the largest departures: blocks with MolmoAct2, where the default reconstruction scores 42\% against 12\% real, and latte with $\pi$0.5, where it scores near zero against 37\% real.}
\label{fig:3}
\end{figure}

Table~\ref{tab:5} and Figure~\ref{fig:3} give the mean score per cell. The authored reconstruction's mean is nearer the real mean in 9 of 10 cells. Three of those margins are 0.50 rubric points or less: blocks with $\pi$0.5 (0.25), bottles with $\pi$0.5 (0.50), and bottles with MolmoAct2, the one cell where the default is nearer (0.30). Differences that small are within the run-to-run spread of a single configuration (Appendix~\ref{app:A}), so we treat them as ties: 7 cells clearly favour the authored reconstruction, 3 are ties, none clearly favours the default reconstruction. The largest margins are stacking bowls with MolmoAct2 (9.00 rubric points), clear table with $\pi$0.5 (5.70), and stacking bowls with $\pi$0.5 and blocks with MolmoAct2 (5.00 each). Under the consistency test of Section~\ref{sec:4}, the authored reconstruction is consistent with the real robot in 9 of 10 cells and the default in 5 of 10. The one cell where the authored reconstruction is not consistent is stacking blocks with MolmoAct2, 7.75 rubric points simulated against 3.50 real; Section~\ref{sec:6} returns to it.

\begin{table}[htbp]
\centering
\small
\setlength{\tabcolsep}{4pt}
\renewcommand{\arraystretch}{1.15}
\caption{Mean score per cell in rubric points; the task maximum is in brackets after the task name, so 9.50 for bottles is 9.50 out of 20. ``consistent'' means the 95\% bootstrap interval on the simulated minus real mean includes zero (Section~\ref{sec:4}). Margin is how much closer to the real mean the authored reconstruction is than the default, in rubric points; a negative margin means the default is nearer.}
\label{tab:5}
\resizebox{\textwidth}{!}{\begin{tabular}{lllllllll}
\toprule
\textbf{task} & \textbf{policy} & \textbf{real} & \textbf{real 95\% CI} & \textbf{authored} & \textbf{authored consistent} & \textbf{default} & \textbf{default consistent} & \textbf{margin} \\
\midrule
bottles in bin (20) & MolmoAct2 & 9.50 & 7.15 to 11.75 & 5.95 & yes & 6.25 & yes & -0.30 \\
bottles in bin (20) & $\pi$0.5 & 10.05 & 7.65 to 12.35 & 9.40 & yes & 8.90 & yes & 0.50 \\
stack bowls (50) & MolmoAct2 & 29.75 & 23.50 to 35.75 & 25.75 & yes & 16.75 & no & 9.00 \\
stack bowls (50) & $\pi$0.5 & 29.25 & 23.00 to 35.50 & 30.50 & yes & 23.00 & yes & 5.00 \\
stack blocks (30) & MolmoAct2 & 3.50 & 1.50 to 5.50 & 7.75 & no & 12.75 & no & 5.00 \\
stack blocks (30) & $\pi$0.5 & 6.00 & 4.00 to 8.25 & 5.75 & yes & 5.50 & yes & 0.25 \\
move latte cup (10) & MolmoAct2 & 5.90 & 4.50 to 7.30 & 5.00 & yes & 3.75 & no & 1.25 \\
move latte cup (10) & $\pi$0.5 & 3.65 & 2.25 to 5.10 & 3.75 & yes & 0.25 & no & 3.30 \\
clear table (70) & MolmoAct2 & 45.35 & 34.25 to 55.40 & 42.50 & yes & 38.65 & yes & 3.85 \\
clear table (70) & $\pi$0.5 & 47.80 & 39.85 to 55.50 & 41.40 & yes & 35.70 & no & 5.70 \\
\bottomrule
\end{tabular}}
\end{table}
\FloatBarrier

\FloatBarrier

\begin{figure}[htbp]
\centering
\includegraphics[width=\linewidth]{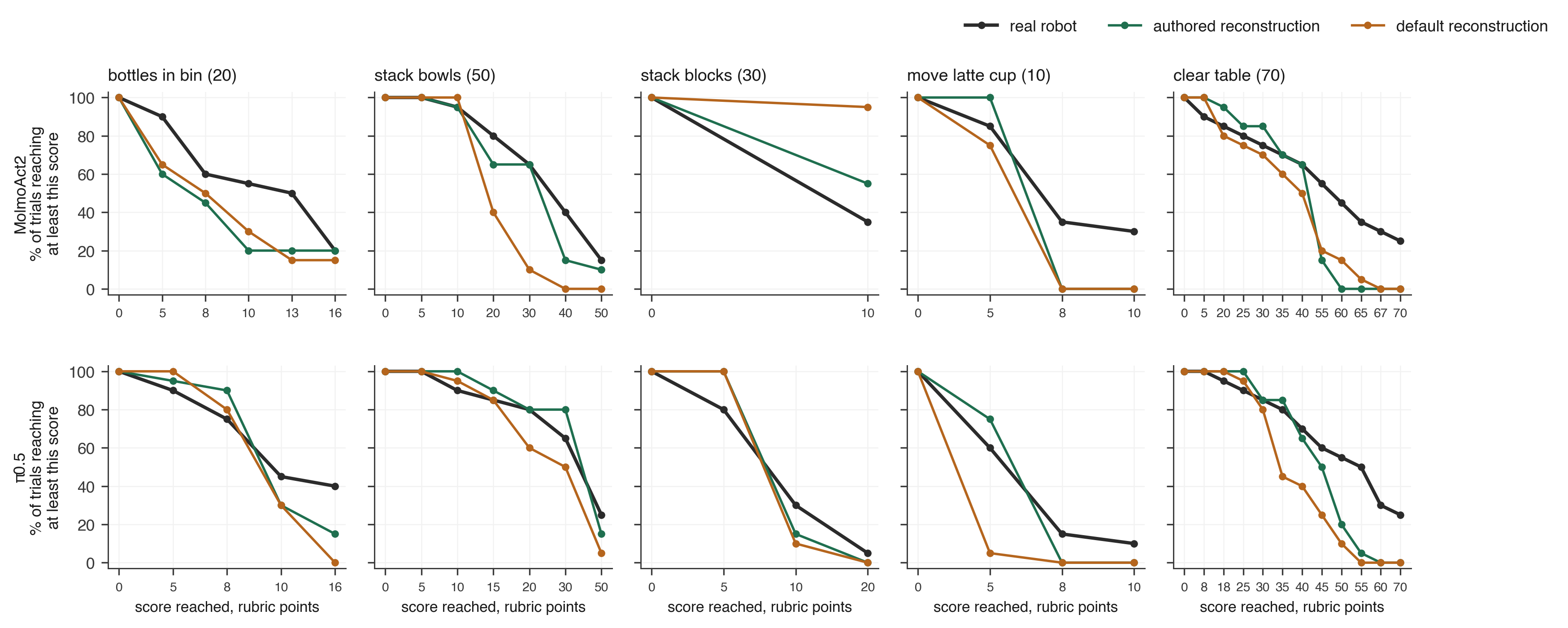}
\caption{Fraction of trials reaching at least each rubric stage, per cell. Stages are the distinct non-zero scores the real robot produced in that cell, so the ladder is the real robot's. A curve that falls earlier is a system that stops earlier. Table~\ref{tab:6} is the mean absolute gap between each simulated curve and the real curve.}
\label{fig:4}
\end{figure}

Table~\ref{tab:6} and Figure~\ref{fig:4} give progress disagreement per cell. The authored reconstruction is closer for progress disagreement on 7 of 10 cells; the largest margins are blocks with MolmoAct2, 20.0 against 60.0 percentage points, and stacking bowls with MolmoAct2, 7.5 against 25.8. Blocks with MolmoAct2 has only one rubric stage above zero in the real data, so its progress value is computed from a single stage. Excluding that cell, the paired difference on progress disagreement is 4.80 percentage points (0.73 to 9.44), so the interval still excludes zero.

\begin{table}[htbp]
\centering
\small
\setlength{\tabcolsep}{4pt}
\renewcommand{\arraystretch}{1.15}
\caption{Progress disagreement per cell, the mean absolute gap in percentage points between the fractions of simulated and of real trials reaching each rubric stage. Lower is closer. ``stages'' is the number of rubric stages the real robot reached.}
\label{tab:6}
\begin{tabular}{lrrr}
\toprule
\textbf{cell} & \textbf{authored} & \textbf{default} & \textbf{stages} \\
\midrule
bottles, MolmoAct2 & 22.0 & 20.0 & 5 \\
bottles, $\pi$0.5 & 15.0 & 17.5 & 4 \\
bowls, MolmoAct2 & 7.5 & 25.8 & 6 \\
bowls, $\pi$0.5 & 6.7 & 10.0 & 6 \\
blocks, MolmoAct2 & 20.0 & 60.0 & 1 \\
blocks, $\pi$0.5 & 13.3 & 15.0 & 3 \\
latte, MolmoAct2 & 26.7 & 25.0 & 3 \\
latte, $\pi$0.5 & 13.3 & 26.7 & 3 \\
clear table, MolmoAct2 & 19.1 & 18.2 & 11 \\
clear table, $\pi$0.5 & 15.5 & 24.1 & 11 \\
\bottomrule
\end{tabular}
\end{table}
\FloatBarrier

\FloatBarrier

\FloatBarrier
\section{Discussion}\label{sec:6}
The results show that reconstructing the objects and scene with metrically scaled geometry and authored physics reduces score error by 10.56 percentage points and progress disagreement by 8.32 percentage points relative to the PolaRiS-inspired recipe. However, there are five limitations to this study.

First, the authored and default reconstructions differ in object geometry, texture, metric scale, collision geometry, mass, friction, restitution and scene at once, so the current design attributes the reduction to the pipeline as a whole and cannot say how much comes from the objects, how much from the scene, or which object property matters most. As an extension of this work, we would change one factor at a time from the default reconstruction, geometry, scale, physics or scene, and measure the change in score error after each.

Second, one cell fails under both reconstructions. Stacking blocks with MolmoAct2 is the one cell where the authored reconstruction is inconsistent with real; the default is inconsistent on that cell too, and both over-score, which points to a cause the two reconstructions share. The real robot scored zero on 13 of 20 trials \cite{ref11}, a mean of 3.50 rubric points; no simulated trial in either reconstruction scored zero, and the means are 7.75 for the authored reconstruction and 12.75 for the default. Development runs on this cell included varying finger-pad friction from 0.8 to 1.4, camera exposure and controller gains but did not close this gap: no setting with two or more trials had a mean below 5 points. The likeliest shared cause is the control gap (Section~\ref{sec:1}): the simulated arm, policy and its controller are identical in both reconstructions, so any difference between how the simulated arm and the real arm respond to the same command adds to both reconstructions' error equally. The cell is kept in every table and figure.

Third, simulated and real trials are graded by different graders, which affects both reconstructions equally. Real trials are scored by the evaluator's graders against the rubric in Table~\ref{tab:3}, while simulated trials are scored automatically by rules on object poses and gripper contact derived from that rubric. Where a rule and a person would disagree, for example a bottle tilted just under the rule's upright threshold, which the rule counts as upright but a grader might not, the disagreement is currently charged to the reconstruction rather than to the scoring. We graded a sample of 30 simulated trials by eye, three per policy and task from the reported batches, and agreed with the rules on 28 of them; the two differences were each one rubric stage and went in opposite directions.

Fourth, at five tasks by two policies the study is small for this area: PolaRiS evaluates six environments by four policies \cite{ref2}, SimFoundry seven tasks by five policies \cite{ref3}, and Wang et al. nine by five \cite{ref4}. We omit MMRV, whether a simulator preserves the real ranking of policies, because with two policies it rests on the real robot's per-task ordering, which is confident on only two of the five tasks (Appendix~\ref{app:B}); with ten cells the intervals on correlation and failure-stage disagreement are too wide to separate the reconstructions (Table~\ref{tab:4}). Bottles with MolmoAct2 shows the variance of twenty-trial batches clearly: its three authored batches scored 3.45, 5.95 and 6.50 against a real mean of 9.50, and it is the one cell where the default is nearer, by 0.30 rubric points. Object placement in simulation varies by at most 1 cm per trial while real placement varies more between trials, and trials are paired by cell rather than one to one, so part of each reconstruction's remaining gap to real is placement, not only reconstruction. Future work should expand the number of tasks, policies and trials per cell.

Finally, two choices in the setup could favour the authored reconstruction over the default. Four of the sixteen object kinds are concave (bowl, bin, cup, mug), and concave shapes are where the default's convex-decomposition colliders depart most from the true surface, filling the bowl and the mug (Figure~\ref{fig:1}). The default reconstruction was built by us following the published PolaRiS recipe, with one photograph per object where PolaRiS uses several and with metric scale set by hand, a step the recipe leaves to the user.

\FloatBarrier
\section{Conclusion}\label{sec:7}
Simulated evaluation tracks the real robot more closely when the objects and scene are built with metrically scaled geometry and authored physics than when they are built with the default open-source recipe. On one evaluator-operated robot setup with two policies and five tasks, the authored reconstruction reduces the gap to the real robot: mean score error is 6.97 percentage points for the authored reconstruction and 17.54 for the default, a reduction of 10.56 percentage points (95\% CI 5.03 to 17.11), and progress disagreement is 15.90 percentage points for the authored reconstruction and 24.23 for the default, a reduction of 8.32 percentage points (1.78 to 16.74). The authored reconstruction is consistent with the real robot's score in 9 of 10 cells against 5 of 10 for the default, and correlates with real cell means at r = 0.90 against 0.51 for the default. Since the two reconstructions differ in geometry, scale, physics and scene at once, this is the effect of the pipeline as a whole; the effect of the individual factors remains to be established. We release the harness, the paired per-trial scores, every reported run's configuration, and the assets and scenes of both reconstructions.

\appendix
\titleformat{\section}[hang]{\normalfont\Large\bfseries}{Appendix \thesection}{1em}{}
\FloatBarrier
\section{Provenance and batch selection}\label{app:A}
\begin{figure}[htbp]
\centering
\includegraphics[width=\linewidth]{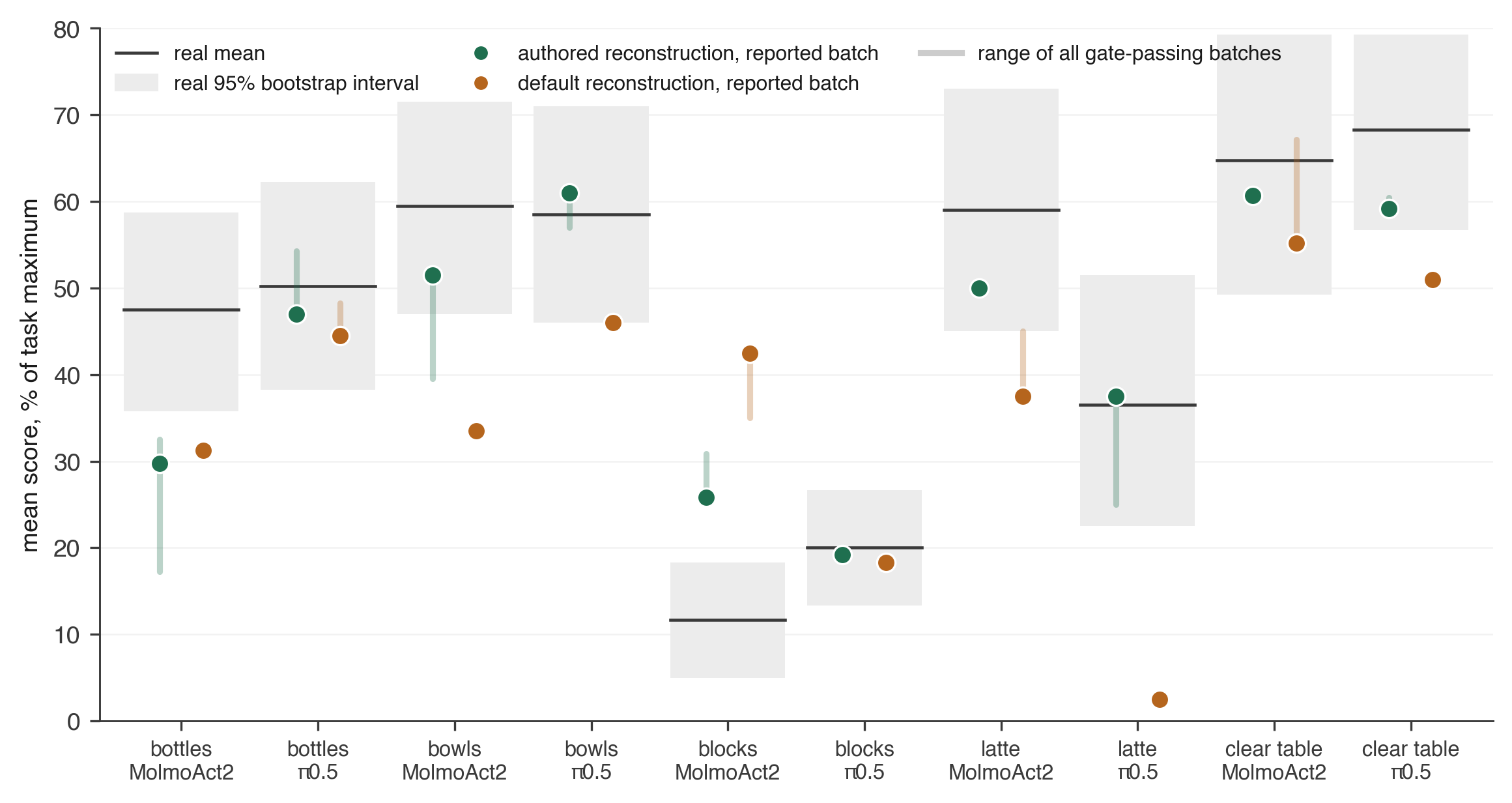}
\caption{Per cell, in percent of task maximum: the real robot's mean (line) and its 95\% bootstrap interval (grey band); the reported batch for the authored (green) and default (orange) reconstruction; and, as a faint bar behind each marker, the range spanned by every batch that passed the configuration gate for that cell and reconstruction. A cell with no bar had a single passing batch. A marker inside the grey band is not the consistency test used in the text, which compares the two means with their combined uncertainty (Section~\ref{sec:4}). The batches themselves are listed in the table above.}
\label{fig:5}
\end{figure}

Every run id is a directory holding results.jsonl and run.log, plus job.json for MolmoAct2 runs. The reported set is committed as scoring/benchmark\_runs.txt. Where more than one batch passed the gate for a cell and reconstruction, the reported batch is the most recent by the run's own logged start time. Every passing batch is listed in Table~\ref{tab:7} and plotted in Figure~\ref{fig:5}. Batch spread within one configuration is 0.00 to 8.35 rubric points, median 1.45, across the 13 cell-reconstruction pairs with more than one batch. 

\begin{table}[htbp]
\centering
\small
\setlength{\tabcolsep}{4pt}
\renewcommand{\arraystretch}{1.15}
\caption{Every gate-passing batch per cell and reconstruction, mean score in rubric points with the run id in brackets; REPORTED marks the batch used in Tables 4 to 6.}
\label{tab:7}
\begin{tabularx}{\textwidth}{>{\raggedright\arraybackslash}X>{\raggedright\arraybackslash}X>{\raggedright\arraybackslash}X}
\toprule
\textbf{cell} & \textbf{authored, all gate-passing batches} & \textbf{default, all gate-passing batches} \\
\midrule
bottles, MolmoAct2 & 3.45 (V2\_K\_bottles\_molmo), REPORTED 5.95 (K2\_bot\_molmo\_plain), 6.50 (A1\_bot\_molmo\_plain) & REPORTED 6.25 (S1\_PS\_bot\_molmo) \\
bottles, $\pi$0.5 & REPORTED 9.40 (K1\_bot\_pi05\_plain), 10.80 (FINAL\_K\_bottles\_pi05), 10.85 (A3\_bot\_pi05\_plain) & REPORTED 8.90 (B0\_PS\_bot\_pi05\_r2), 9.50 (S6\_PS\_bot\_pi05), 9.65 (A0\_PS\_bot\_pi05\_rerun) \\
bowls, MolmoAct2 & 19.75 (six jitter-sweep shards pooled: 04\_NS10, JIT25, JIT40, DBG\_JIT25, DBG\_JIT40, PROBE\_NS10; the gate cannot see the placement shift, so they pass it; listed, not reported), 20.00 (N20\_K\_bowls\_molmo), 23.25 (V2\_K\_bowls\_molmo), REPORTED 25.75 (RB\_BOWLS\_a + RB\_BOWLS\_b) & REPORTED 16.75 (S2\_PS\_bowls\_molmo) \\
bowls, $\pi$0.5 & 28.50 (FINAL\_K\_bowls\_pi05), REPORTED 30.50 (C2\_bowls\_pi05) & REPORTED 23.00 (S7\_PS\_bowls\_pi05) \\
blocks, MolmoAct2 & REPORTED 7.75 (C1\_blocks\_molmo), 8.00 (N20\_K\_blocks\_molmo), 9.25 (V2\_K\_blocks\_molmo) & 10.50 (N20\_P\_blocks\_molmo), 12.50 (V2\_P\_blocks\_molmo), REPORTED 12.75 (S3\_PS\_blocks\_molmo) \\
blocks, $\pi$0.5 & REPORTED 5.75 (T1\_blocks\_pi05), 6.00 (FINAL\_K\_blocks\_pi05) & REPORTED 5.50 (ZZ1\_PS\_blocks\_pi05) \\
latte, MolmoAct2 & 5.00 (N20\_K\_latte\_molmo), REPORTED 5.00 (V2\_K\_latte\_molmo) & REPORTED 3.75 (S4\_PS\_latte\_molmo), 4.50 (V2\_P\_latte\_molmo) \\
latte, $\pi$0.5 & 2.50 (FINAL\_K\_latte\_pi05), REPORTED 3.75 (K3\_latte\_pi05\_plain) & REPORTED 0.25 (S9\_PS\_latte\_pi05) \\
clear table, MolmoAct2 & REPORTED 42.50 (V2\_K\_cleartable\_molmo) & REPORTED 38.65 (S5\_PS\_ct\_molmo), 47.00 (V2\_P\_cleartable\_molmo) \\
clear table, $\pi$0.5 & REPORTED 41.40 (ZZ2\_K\_ct\_pi05), 42.30 (FINAL\_K\_cleartable\_pi05) & REPORTED 35.70 (SA\_PS\_ct\_pi05) \\
\bottomrule
\end{tabularx}
\end{table}
\FloatBarrier

\FloatBarrier

\FloatBarrier
\section{Policy ranking}\label{app:B}
Ranking preservation and MMRV \cite{ref1} need the real reference to order the policies. The bootstrap probability that MolmoAct2 outscores $\pi$0.5 on the real robot is 0.36 for bottles, 0.53 for bowls, 0.04 for blocks, 0.98 for latte and 0.36 for clear table. Two of the five tasks are ordered with confidence; both reconstructions agree with real on one of them and disagree on the other. We do not report MMRV on this basis.

\FloatBarrier
\section{Reproduction}\label{app:C}
\path{build_table.py} \texttt{--runs} \texttt{RUNS} \texttt{--real} \path{real_reference_pertrial.csv} \texttt{--runs-list} \path{benchmark_runs.txt} \texttt{--json-out} \path{table.json} builds Table~\ref{tab:5} from the committed run list. \path{agreement.py} \path{table.json} \texttt{--real} \path{real_reference_pertrial.csv} produces Tables~\ref{tab:4} and~\ref{tab:6}; \path{median_table.py} \path{table.json} is the median check. Figures~\ref{fig:3} to~\ref{fig:5} are produced by the scripts in \path{figures/} from the same JSON. Both scoring scripts default to bootstrap seed 0, which reproduces every interval in Tables 4 to 6. The scoring scripts and the trial runner are released with the paper at \href{https://github.com/Kaedim/yam-sim-harness-oss}{github.com/Kaedim/yam-sim-harness-oss}, commit d3817f7; the runner that produced the reported runs has md5 3274dcbf. The run artefacts, the real per-trial scores, the assets and scenes of both reconstructions are released as Hugging Face datasets, linked from the repository README.

\section*{Funding and competing interests}
The real-robot trials were commissioned and paid for by Kaedim under a benchmarking agreement with Robocurve. Kaedim specified the three custom tasks and agreed the policies with Robocurve; Robocurve designed the rubrics, ran and graded the real trials, and under the agreement retained control of real-world methodology, scoring and interpretation. The authored reconstruction's objects were produced with Kaedim's asset pipeline (\href{https://www.projectsim.ai}{projectsim}) and its scene with Kaedim's scene reconstruction pipeline; the authors are Kaedim employees. Naming Robocurve does not imply its endorsement of the results.


\clearpage
\begin{thebibliography}{99}
\bibitem{ref1} Li et al. Evaluating Real-World Robot Manipulation Policies in Simulation (SIMPLER). \href{https://arxiv.org/abs/2405.05941}{arXiv:2405.05941}, 2024.
\bibitem{ref2} Jain et al. PolaRiS: Scalable Real-to-Sim Evaluations for Generalist Robot Policies. \href{https://arxiv.org/abs/2512.16881}{arXiv:2512.16881}, 2025.
\bibitem{ref3} Ranawaka et al. SimFoundry: Modular and Automated Scene Generation for Policy Learning and Evaluation. \href{https://arxiv.org/abs/2606.28276}{arXiv:2606.28276}, 2026.
\bibitem{ref4} Wang et al. A Practical Recipe Towards Improving Sim-and-Real Correlation for VLA Evaluation. \href{https://arxiv.org/abs/2606.10366}{arXiv:2606.10366}, 2026.
\bibitem{ref5} Wang et al. Phys2Real: Fusing VLM Priors with Interactive Online Adaptation for Uncertainty-Aware Sim-to-Real Manipulation. \href{https://arxiv.org/abs/2510.11689}{arXiv:2510.11689}, 2025.
\bibitem{ref6} Zhang et al. Real-to-Sim Robot Policy Evaluation with Gaussian Splatting Simulation of Soft-Body Interactions. \href{https://arxiv.org/abs/2511.04665}{arXiv:2511.04665}, 2025.
\bibitem{ref7} Fan et al. TwinAligner: Visual-Dynamic Alignment Empowers Physics-aware Real2Sim2Real for Robotic Manipulation. \href{https://arxiv.org/abs/2512.19390}{arXiv:2512.19390}, 2025.
\bibitem{ref8} Jiang et al. PhysTwin: Physics-Informed Reconstruction and Simulation of Deformable Objects from Videos. \href{https://arxiv.org/abs/2503.17973}{arXiv:2503.17973}, 2025.
\bibitem{ref9} Xiang et al. Structured 3D Latents for Scalable and Versatile 3D Generation (TRELLIS). \href{https://arxiv.org/abs/2412.01506}{arXiv:2412.01506}, 2024.
\bibitem{ref10} I2RT Robotics, YAM bimanual arm. \href{https://github.com/i2rt-robotics}{github.com/i2rt-robotics}.
\bibitem{ref11} Robocurve. Real-world evaluation results for Kaedim custom tasks, per-trial results page. \href{https://kaedim.pages.dev/real-world-evals/}{kaedim.pages.dev/real-world-evals}, 2026. Source of all real scores, rubrics, camera heights and step limits.
\bibitem{ref12} Physical Intelligence. $\pi$0.5: a Vision-Language-Action Model with Open-World Generalization. \href{https://arxiv.org/abs/2504.16054}{arXiv:2504.16054}, 2025.
\bibitem{ref13} Allen Institute for AI. MolmoAct 2, technical report. \href{https://allenai.org/papers/molmoact2}{allenai.org/papers/molmoact2}, 2026.
\bibitem{ref14} Kerbl et al. 3D Gaussian Splatting for Real-Time Radiance Field Rendering. \href{https://arxiv.org/abs/2308.04079}{arXiv:2308.04079}, 2023.
\bibitem{ref15} Wei et al. Approximate Convex Decomposition for 3D Meshes with Collision-Aware Concavity and Tree Search (CoACD). \href{https://arxiv.org/abs/2205.02961}{arXiv:2205.02961}, 2022.
\bibitem{ref16} Huang et al. 2D Gaussian Splatting for Geometrically Accurate Radiance Fields. \href{https://arxiv.org/abs/2403.17888}{arXiv:2403.17888}, 2024.
\bibitem{ref17} Schönberger and Frahm. Structure-from-Motion Revisited (COLMAP). CVPR, 2016. \href{https://colmap.github.io/}{colmap.github.io}.
\end{thebibliography}
\end{document}